\documentclass{easychair}
\usepackage{paralist}
\usepackage{amssymb,amsmath}
\usepackage{booktabs} 
\title{Self-Learned Formula Synthesis in Set Theory\thanks{Supported by the 
ERC Consolidator grant no. 649043 AI4REASON}}
\author{Chad E. Brown \and  Thibault Gauthier}
\institute{Czech Technical University, Prague}
\authorrunning{Brown, Gauthier}
\titlerunning{Self-Learned Formula Synthesis in Set Theory}

\def\limplies{\Rightarrow}
\usepackage{tikz} 
\usetikzlibrary{shapes,arrows}
\usepackage{pgfplots}
\pgfplotsset{compat=newest}

\begin{document}
\maketitle

One of the most difficult tasks in higher-order theorem proving
is the instantiation of set variables~\cite{BledsoeF93,Brown02}.
An important class of theorem proving problems requiring instantiation
of a set variable are those requiring induction~\cite{hipspec13}.
Instantiating a set variable often requires synthesizing a formula
satisfying some properties. In our work we apply machine
learning to the task of synthesizing formulas satisfying
a collection of semantic properties. Previous work applying machine learning to
induction theorem proving can be found in~\cite{DBLP:conf/aisc/JiangPF18}.

\paragraph{Hereditarily finite sets}

In~\cite{Ackermann1937} Ackermann proved consistency of Zermelo's axioms of set theory without
an axiom of infinity by interpeting natural numbers $0,1,2,\ldots$ as sets.
Membership $m\in n$ is taken to hold if bit $m$ is $1$ in the binary representation of $n$,
e.g., $0\in 1$, $1\in 2$ and $0\notin 2$. This is known as the {\emph{Ackermann encoding}} of
hereditarily finite sets. We will always consider terms and formulas to be interpreted via the
model given by this encoding.

As terms $s,t$ we take variables $x,y,z,\ldots$ as well as $\wp(t)$ (power set of $t$), $\{t\}$, and $s\cup t$.
As atomic formulas we take $s\in t$, $s\not\in t$, $s\subseteq t$, 
$s\not\subseteq t$, $s=t$ and $s\neq t$.
Formulas $\varphi, \psi$ are either atomic formulas or of the form $\varphi 
\limplies \psi$,
$\varphi\land \psi$, $\forall x\in s.\varphi$, $\exists x\in s.\varphi$,
$\forall x\subseteq s.\varphi$ or $\exists x\subseteq s.\varphi$.
Note that all our quantifiers are bounded. As a consequence, for every assignment of free variables
to natural numbers we can always (in principle) calculate the truth value for a formula under the assignment.
In practice if certain bounds are exceeded evaluation fails.

\paragraph{Formula Generation}
All formulas up to size 15 with at most one free variable $x$ were generated.
For each of these formulas we attempted to evaluate the formula with $x$ 
assigned to values between $0$ and $63$. We call this list of truth values
the \textit{graph} of the formula.
We omitted each formula that failed to evaluate on any of these values.
For the remaining formulas, we kept one representative formula (of minimal size) for each subset of $\{0,\ldots,63\}$
that resulted from an evaluation. This resulted in a set $\mathbb{F}$ of 6750 
formulas varying in size from 3 to 15
distributed as indicated in Table~\ref{tab:size}.

\begin{table}
\begin{tabular}{rccccccccccccc}
  \toprule
  Size & 3 & 4 & 5 & 6 & 7 & 8 & 9 & 10 & 11 & 12 & 13 & 14 & 15\\
  No. of formulas & 6 & 8 & 22 & 60 & 88 & 260 & 472 & 960 & 638 & 992 & 1582 & 
  1056 & 606\\
  \bottomrule
\end{tabular}
\caption{Number of generated formulas of each size}\label{tab:size}
\end{table}

\paragraph{The Formula Synthesis Problem}
The goal of the synthesis task is to create a formula 
with one free variable for a given graph. To ensure that the task can be 
achieved, we choose the graphs of the generated formulas as inputs to our 
problems. For each formula $\varphi \in \mathbb{F}$ the associated problem is 
to find a formula $\psi$ that has the same graph as $\varphi$ by only 
observing the graph of $\varphi$. 
We restrict ourselves to solutions that construct 
$\psi$ from left to right if represented in prefix notation. 

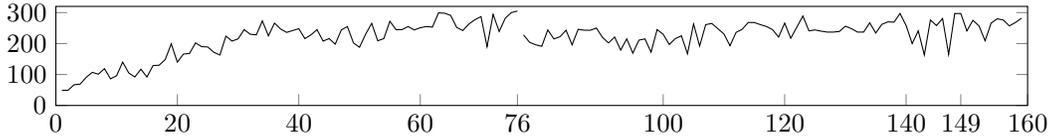
\begin{figure}[ht]
\centering
\begin{tikzpicture}[]
\begin{axis}[
  width=\textwidth,
  height=0.2*\textwidth,
  xmin=0, xmax=160,
  ymin=0, ymax=320,
  xtick pos=left,
  xtick={0,20,40,60,76,100,120,140,149,160},
  ytick={}
  ]
\addplot[color=black] table[x=generation, y=score] {test};
\addplot[color=black] table[x=gen, y=win] {latestrun2};
\end{axis}
\end{tikzpicture}
\caption{\label{fig:run} 
Number of successful formula synthesis (y) at generation (x)}
\vspace{-2mm}
\end{figure}

\paragraph{A Solution by Reinforcement Learning}
Our reinforcement learning framework~\cite{DBLP:journals/corr/abs-1910-11797} 
relies on a curriculum learning approach.
It perfects its synthesis abilities on the easiest problems first before moving
to harder ones. The difficulty of a problem derived from a formula $\varphi 
\in  \mathbb{F}$ is defined to be the size of $\varphi$.
Each level consists of $400$ graphs with lower levels containing easier 
problems.
Each \textit{generation} consists of an exploration phase and a training phase.

During the exploration phase, the algorithm attempts to find a solution for
400 graphs taken in equal measure from each level lower or equal to the current 
level.
An attempt for a graph $g$ consists of a series of big steps. The number of big 
steps is limited to twice the size of $\varphi$.
One big step consists of one call to Monte Carlo tree search algorithm with
a partial formula $\psi$ as the root of the search tree.
The number of search steps for one MCTS call is set at 50000.
Then, the step from the root with the highest number of visits is chosen.
This adds one operator to $\psi$. The updated formula becomes the root of the 
search in the next big step.
The algorithm moves to the next level when it solves strictly more than 75\% 
of the problems in one phase.

During the training phase, a tree neural network (TNN) that predicts both the 
value and the policy is trained on the 200000 newest examples. 
Each of those examples is extracted from the root tree statistics after one big 
step.
Since we perform searches for many different graphs, the information 
about the targeted graph $g$ is given to our network in addition to the 
partially constructed formula $\psi$. They are represented together in the 
tree structure by $\mathit{concat}(\mathit{g'},\psi)$ where 
$g' \in \mathbb{R}^{64}$ is the embedding of $g$ and 
$\mathit{concat}$ is an additional helper operator.
When guiding the MCTS algorithm, noise is added to the 
predicted policy to favor exploration.

\paragraph{Results}
In Figure~\ref{fig:run}, the success rate at each generation of the 
reinforcement learning run
is shown.  Level 1 is passed 
at generation 76 with 305 formulas synthesized. The run is stopped at 
generation 159. In Table~\ref{tab:test}, the TNN from generation 149 is 
tested 
without noise (Guided) on 
problems from level 1, 2 and 3. To produce a baseline, we replace the MCTS 
algorithm by 
a breadth-first search algorithm (Breadth-first).
We also try to figure how much the input graph 
influences the search by masking its embedding $g'$ (Hidden-graph).

\begin{table}[h]
\centering
\begin{tabular}{cccc}
\toprule 
& Breadth-first & Hidden-graph & Guided\\
\cmidrule(lr){2-2} \cmidrule(lr){3-3} \cmidrule(lr){4-4}
Level 1, 2, 3 & 68, 0, 0 & 270, 126, 59 & 338, 240, 165 \\
\bottomrule
\end{tabular}
\caption{Number of successful formula synthesis in level 1, 2 and 3 
respectively}\label{tab:test}
\vspace{-2mm}
\end{table}

The formula $\varphi = \exists y \in x.\ x 
\not\subseteq \wp(y) \in \mathbb{F}$ does not seem to have an obvious meaning. 
From 
the 
graph of 
$\varphi$, the equivalent 
formula  $\psi = \exists y \in x. \ \lbrace y \rbrace \neq x$ is synthesized by 
our algorithm. This reveals that the formula defines the 
predicate for $x$ 
having at least two elements.

\paragraph{Conclusion}
This work indicates that formula synthesis for an assignment of truth 
values can be learned progressively using only guided exploration as an 
improvement mechanism. 
In the 
future, we consider improving the techniques developed and integrating them in 
automated theorem provers~\cite{Brown12,Sch02-AICOMM}.

\bibliographystyle{plain}
\bibliography{ate11}

\begin{thebibliography}{1}

\bibitem{Ackermann1937}
Wilhelm Ackermann.
\newblock Die {W}iderspruchsfreiheit der allgemeinen {M}engenlehre.
\newblock {\em Mathematische Annalen}, 114(1):305--315, 1937.

\bibitem{BledsoeF93}
W.~W. Bledsoe and Guohui Feng.
\newblock Set-var.
\newblock {\em Journal of Automated Reasoning}, 11(3):293--314, 1993.

\bibitem{Brown02}
Chad~E. Brown.
\newblock Solving for set variables in higher-order theorem proving.
\newblock In Andrei Voronkov, editor, {\em Automated Deduction - CADE-18, 18th
  International Conference on Automated Deduction, Copenhagen, Denmark, July
  27-30, 2002, Proceedings}, volume 2392 of {\em Lecture Notes in Computer
  Science}, pages 408--422. Springer, 2002.

\bibitem{Brown12}
Chad~E. Brown.
\newblock Satallax: An automatic higher-order prover.
\newblock In Bernhard Gramlich, Dale Miller, and Uli Sattler, editors, {\em
  IJCAR}, volume 7364 of {\em LNCS}, pages 111--117. Springer, 2012.

\bibitem{hipspec13}
Koen Claessen, Moa Johansson, Dan Ros{\'e}n, and Nicholas Smallbone.
\newblock Automating inductive proofs using theory exploration.
\newblock In Maria~Paola Bonacina, editor, {\em Conference on Automated
  Deduction (CADE)}, volume 7898 of {\em LNCS}, pages 392--406. Springer, 2013.

\bibitem{DBLP:journals/corr/abs-1910-11797}
Thibault Gauthier.
\newblock Deep reinforcement learning in {HOL4}.
\newblock {\em CoRR}, abs/1910.11797, 2019.

\bibitem{DBLP:conf/aisc/JiangPF18}
Yaqing Jiang, Petros Papapanagiotou, and Jacques~D. Fleuriot.
\newblock Machine learning for inductive theorem proving.
\newblock In Jacques~D. Fleuriot, Dongming Wang, and Jacques Calmet, editors,
  {\em Artificial Intelligence and Symbolic Computation - 13th International
  Conference, {AISC} 2018, Suzhou, China, September 16-19, 2018, Proceedings},
  volume 11110 of {\em Lecture Notes in Computer Science}, pages 87--103.
  Springer, 2018.

\bibitem{Sch02-AICOMM}
Stephan Schulz.
\newblock {E - A Brainiac Theorem Prover}.
\newblock {\em AI Commun.}, 15(2-3):111--126, 2002.

\end{thebibliography}
\end{document}